%% file: main.tex
\def\year{2020}\relax
\documentclass[letterpaper]{article} % DO NOT CHANGE THIS
\usepackage{aaai20}  % DO NOT CHANGE THIS
\usepackage{times}  % DO NOT CHANGE THIS
\usepackage{helvet} % DO NOT CHANGE THIS
\usepackage{courier}  % DO NOT CHANGE THIS
\usepackage[hyphens]{url}  % DO NOT CHANGE THIS
\usepackage{graphicx} % DO NOT CHANGE THIS
\usepackage{graphicx}  % DO NOT CHANGE THIS
\nocopyright

\usepackage{amsmath, amssymb, amsbsy}
\usepackage{booktabs, multirow}
\newcommand{\citet}[1]{\citeauthor{#1} \shortcite{#1}} 
\newcommand{\citep}{\cite}

\DeclareMathOperator*{\argmin}{argmin}

\newcommand{\cut}[1]{}
\newcommand{\mbf}[1]{\mathbf{#1}}

\title{Integrating Physiological Time Series and Clinical Notes \\with
Deep Learning for Improved ICU Mortality Prediction}
\author{Satya Narayan Shukla, Benjamin M. Marlin\\
College of Information and Computer Sciences\\
University of Massachusetts Amherst\\
\texttt{\{snshukla, marlin\}@cs.umass.edu}
}

\begin{document}

\maketitle

\begin{abstract}
Intensive Care Unit Electronic Health Records (ICU EHRs) store multimodal data about patients including clinical notes, sparse and irregularly sampled physiological time series, lab results, and more. To date, most methods designed to learn predictive models from ICU EHR data have focused on a single modality. In this paper, we leverage the recently proposed interpolation-prediction deep learning architecture \citep{shukla2019interpolationprediction} as a basis for exploring how physiological time series data and clinical notes can be integrated into a unified mortality prediction model. We study both early and late fusion approaches, and demonstrate how the relative predictive value of clinical text and physiological data change over time. Our results show that a late fusion approach can provide a statistically significant improvement in mortality prediction performance over using individual modalities in isolation.
\end{abstract}

\input{intro.tex}
\input{related.tex}

\input{model.tex}

\input{experiments.tex}

\input{conclusions.tex}

\bibliographystyle{aaai}
\bibliography{amia}

\end{document}

%% file: intro.tex
%!TEX root = amia.tex

\section{Introduction}
Electronic health records (EHRs) store multimodal data related to individual medical history including clinical notes, physiological measurements, lab results, radiology images, and more. Intensive Care Unit Electronic Health Records (ICU EHRs) are particularly interesting as they contain measurements of multiple physiological variables through time. The analysis of these data has the potential to improve care via the creation of improved decision support tools based on machine learning and data mining techniques. However, the complexity of the data has led to a focus on analyzing single data modalities in isolation \citep{marlin-ihi2012,lipton2016directly,che2016recurrent,futoma2017improved}. 

In this paper, we explore the predictive value of integrating physiological time series data and clinical text into a unified mortality prediction model. Specifically,  we leverage the content of clinical notes through time and fuse the information they contain with physiological time series data. We build on the recently proposed  interpolation-prediction deep learning architecture as a framework for modeling sparse and irregularly sampled physiological time series data \citep{shukla2019interpolationprediction}. We study several methods for representing the clinical text, along with both early and late fusion approaches to integrating the two data modalities \citep{kiela,shortfuse}. 

We begin by presenting related work on physiological time series modeling, clinical text, and fusion approaches. We next present the proposed approach including a brief review of interpolation-prediction networks. Finally, we present mortality prediction experiments on the MIMIC-III data set \citep{johnson2016mimic} demonstrating how the relative predictive value of clinical text and physiological data change during the first 48 hours after admission. We show that the late fusion approach can provide a significant improvement over using individual modalities in isolation.

%% file: related.tex
%!TEX root = amia.tex

\section{Related Work}
\label{sec:related}

The problem of interest in this work is learning supervised 
machine learning models by fusing  clinical time series with unstructured clinical text data. In this section, we review related work on modeling and analysis of both clinical text and sparse and irregularly sampled physiological time series, as well as related work on fusion approaches.

\subsection{Clinical text}
With increasing access to clinical notes over the last several years, there has been significant progress in understanding clinical text data and using these data to improve prediction of clinical outcomes. Natural language processing and information extraction techniques have been successfully applied to tasks including  clinical concept extraction \citep{cliner}, relation extraction \citep{jamia2}, question-answering \citep{jamia3}, predictive modeling \citep{ghasemmi} and more.  

Methods for using narrative notes to predict clinical outcomes include medical concept extraction using rule based \citep{jamia2} or machine learning techniques \citep{cliner}. However, such methods can require a substantial amount of work in rule construction, keyword selection, text annotation or feature engineering for supervised machine learning. Unsupervised methods such as topic modeling can be used to solve this problem. Topic modeling  \citep{ghasemmi} methods rely on extracting topic features from clinical text data. \citet{lehman} combined both topic modeling and medical concept extraction approaches for predicting in-hospital mortality.  

Inspired by the recent success of word embedding methods including word2vec and GloVe \citep{glove} in numerous natural language processing tasks, \citet{minarro} learned an embedding model for medical text data. \citet{devine} used journal abstracts to train the embeddings. \citet{choiconcept} evaluated the efficiency of word embeddings in capturing relations between medical concepts. \citet{boug} compared the clinical notes representation generated by Bag of Words (BOW), word2vec  and the final hidden layer of a learned LSTM for downstream clinical prediction tasks. Their results showed that there is no simple winning representation. BoW and word2vec achieved similar performance in predicting in-hospital mortality. In other recent work, \citet{ghassemi15} used Gaussian processes to model sequences of clinical notes as time series of topics. 

%\citet{lapata08} and \citet{lapata12} used word embeddings to compute phrase/sentence embedding using operations on vectors and matrices. \citet{leandmikolov} assumed a latent representation for each paragraph which influences the distribution of words in it. Skip-thought \citep{kiros} learns paragraph embeddings by trying to reconstruct the surrounding sentences. RNNs such as LSTM and GRU, which capture long term dependency, have also been used to model sentences \citep{tai15}. \citet{blunsom} used convolutional models for sentence modeling and achieved good results for sentiment prediction and classification tasks. Unweighted \citep{word2vec} as well as weighted \citep{simple} averaging of word embeddings have also been used to compute sentence embeddings. None of the prior work utilize the sentence embedding methods for modeling long clinical notes.

%A separate line of work has looked at directly modeling sequential clinical data without using hand-engineered features. For example, \citet{liptonlearningTD} and \citet{choi} modeled physiological time series and codes using recurrent networks, \citet{razavian16} applied convolutional model to lab results. 

In this work, we consider clinical notes as sequences of words or sentences and use recurrent networks to predict in-hospital mortality. We generate sentence embeddings using simple averages \citep{glove} as well as weighted  averages of word embeddings \citep{simple}. Similar to \citet{boug}, we compare with Bag-of-Words and GloVe models. Finally, similar to \citet{blunsom}, we also use a convolutional model for prediction where the clinical notes are represented in terms of word embeddings.

\subsection{Irregularly sampled physiological time-series}
A sparse and irregularly sampled time series is a sequence of 
samples with large and irregular intervals between their 
observation times. Such data commonly occur in electronic 
health records where they can represent a significant problem 
for both supervised and ussupervised learning methods \citep{marlin-ihi2012}.

A closely related problem is performing supervised learning in the
presence of missing data \citep{little2014statistical}. Indeed,
the problem of analyzing sparse and irregularly sampled data can
be converted into a missing data problem (typically with loss of information
or inference efficiency) by discretizing the time axis and indicating that
intervals  with no observed samples are missing.
This is the approach taken to deal with irregular sampling by \citet{marlin-ihi2012}
as well as \citet{lipton2016directly}. Learning is generally harder as the amount of missing data increases, so choosing a discretization interval length must be 
dealt with as a hyper-parameter of such a method.

The alternative to pre-discretization 
is to construct models with the ability to directly use
an irregularly sampled time series as input. 
For example, \citet{Lu2008} present a kernel-based method that can be used
to produce a similarity function between two irregularly sampled time
series. \citet{li2015classification} subsequently provided a 
generalization of this approach to the case of kernels between Gaussian
process models. \citet{li2016scalable} showed how a deep neural network model
(feed-forward, convolutional, or recurrent)
could instead be stacked on top of a Gaussian process layer with end-to-end training, while \citet{futoma2017improved} showed how this approach could be generalized from the univariate to the multivariate setting.

An important property of the above models is that they allow for incorporating 
all of the information from all available time points into a global interpolation 
model. A separate line of work has looked at the use of more local interpolation
methods while still operating directly over continuous-time inputs.
For example, \citet{che2016recurrent}
presented several methods based on  gated recurrent unit (GRU) networks
\citep{gru} combined with simple imputation methods
including mean imputation and forward filling with past values. 
\citet{che2016recurrent} additionally considered an approach 
that takes as input a sequence consisting of both the
observed values and the timestamps at which those values were observed. 
The previously observed input value is decayed over time 
toward the overall mean. In another variant, the hidden states are 
similarly decayed toward zero. \citet{Yoon_mRNN} presented another similar approach based on a multi-directional RNN, which operates across streams in addition to
within streams.   

In this work, we use the recently proposed interpolation-prediction network to model sparse and irregularly sampled physiological time series \citep{shukla2019interpolationprediction}. This framework 
addresses some difficulties with prior 
approaches including the complexity of the Gaussian process
interpolation layers used in \citet{li2016scalable} and \citet{futoma2017improved}, and 
the lack of modularity in the approach of \citet{che2016recurrent}. We describe this framework in more detail in Section \ref{sec:interp}.
%The interpolation-prediction network achieves state-of-the-art results on mortality prediction task from physiological time series data.

\subsection{Fusion models}
Learning multimodal representations is a fundamental 
research problem. Canonical correlation analysis \citep{harold} has been 
widely used for modeling multimodal data \citep{hardoon,klein}. 
It learns a subspace to maximize the correlation between multiple modalities. 
\citet{ngiam} introduced a multimodal deep learning framework to combine video 
and audio for speech recognition. Multimodal learning with language and vision 
subspaces has been used to improve the performance of image captioning 
tasks \citep{karpathy14,socher14}. 
\citet{srivastavamulti} used fused representations of multiple 
modalities (text and images) as input for discriminative tasks.
\citet{silberer} use stacked autocoders for fusing multimodal 
data while \citet{kiela} adopt a simple concatenation strategy 
and achieve empirical improvements using convolutional 
models for extracting visual features and skip-gram models for text. 

A separate line of work has looked at combining time-series data and textual information. \citet{tang} analyzed news reports to improve the prediction of stock prices, while \citet{rodrigues} use a simple concatenation approach to 
combine time-series and textual data for taxi demand prediction by learning their latent representations.

Within the clinical data space, \citet{shortfuse} showed 
how combining structured 
information like age, gender, height, etc. with time 
series data can improve performance. \citet{xu_multi} developed
clinical predictive models by integrating continuous monitoring data with discrete clinical event sequences.
\citet{rajkomar_multi}
combined multiple modalities such as demographics, provider orders, diagnoses, procedures, medications,
laboratory values, clinical text data and vital signs and showed improved performance on multiple tasks.
 \citet{mlforh} 
combined unstructured clinical text data 
with physiological time-series data for
in-hospital mortality prediction, similar to the present work.
Relative to that work, we consider multiple clinical text 
representations, base our time series model on 
interpolation-prediction networks \citep{shukla2019interpolationprediction},
and focus on how the relative value of clinical text and physiological data vary 
through time. Further, we consider both early and late
fusion approaches, expanding on the prior work of
\citet{kiela} and \citet{shortfuse}.

%% file: model.tex
%!TEX root = amia.tex

\begin{figure*}[t]
\centering
\includegraphics[width=0.9\linewidth]{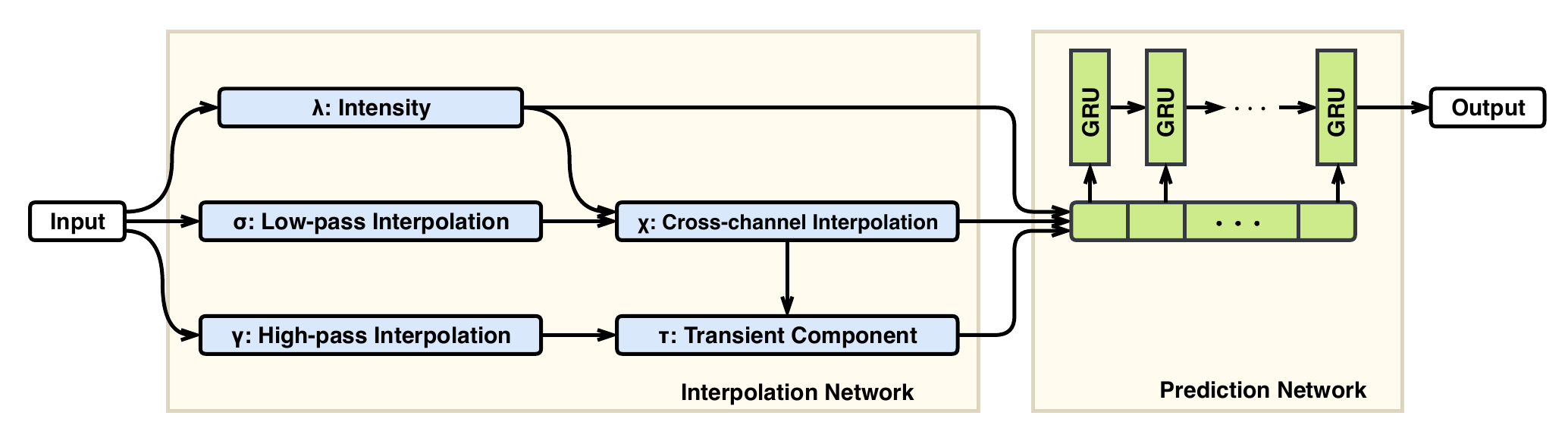}
\caption{Interpolation-Prediction Network}
\label{fig:model_ip}
\end{figure*}

\section{Proposed Fusion Model Framework}
In this section, we present the proposed fusion modeling framework. We begin by presenting notation and a description of the models used for clinical text and physiological time series, followed by a discussion of fusion approaches.

\subsection{Notation}

We let $\mathcal{D}=\{(\mbf{s}_n,\mbf{v}_n,y_n)|n=1,...,N\}$ represent a data set containing
$N$ data cases. An individual data case consists of
a single target value $y_n$ (discrete in the case of classification), a $D$-dimensional, sparse and irregularly
sampled multivariate physiological time series $\mbf{s}_n$, and the unstructured text data present in the clinical notes $\mbf{v}_n$, represented as a sequence of words. We note that different dimensions $d$ of the multivariate
time series $\mbf{s}_n$ can have observations at different times, as well as different total
numbers of observations $L_{dn}$. Thus, we represent time series $d$ for data case $n$ as a tuple
$\mbf{s}_{dn}=(\mbf{t}_{dn}, \mbf{x}_{dn})$  where $\mbf{t}_{dn}=[t_{1dn},...,t_{L_{dn}dn}]$
is the list of time points at which observations are defined and 
$\mbf{x}_{dn}=[x_{1dn},...,x_{L_{dn}dn}]$ is the corresponding list of observed values.

\subsection{Time Series Model}
\label{sec:interp}
We briefly review the interpolation-prediction network framework that \citet{shukla2019interpolationprediction} proposed to model sparse 
and irregularly sampled time series data. The architecture is based on the use of several semi-parametric interpolation 
layers organized into an interpolation network, followed by the application of a prediction network that can leverage 
any standard deep learning model. Figure \ref{fig:model_ip} shows the architecture of the interpolation-prediction model.

The interpolation network interpolates
the multivariate, sparse, and irregularly sampled input time series against 
a set of reference time points $\mbf{r}= [r_1, \cdots, r_T]$. A two-layer interpolation network is used where the first layer separately transforms each of $D$ univariate input time series, creating several intermediate interpolants. The second interpolation layer merges information across all time series at each reference time point by taking into account learnable correlations across all time series. The interpolation network outputs a total of $C=3$ components for each dimension of the input time series: 
a smooth, cross-channel interpolant to capture smooth trends, a transient component to capture transients, and an intensity function to capture information about where observations occur in time. We define $f_\theta(\mbf{s}_n)$ to be the function computing the output $\hat{\mbf{s}}_n$ of the interpolation network. The output  $\hat{\mbf{s}}_n$ is a fixed-sized array with dimensions $(DC) \times T$. 

The second component, the prediction network, takes the output of the interpolation network $\hat{\mbf{s}}_n$ as its input and produces a prediction $\hat{y}_n=g_{\omega}(\hat{\mbf{s}}_n)=
g_{\omega}(f_{\theta}(\mbf{s}_n))$ for the target variable $y_n$. The prediction network can consist of any standard supervised neural network architecture (fully-connected feedforward, convolutional, recurrent, etc). 

We learn the parameters of this model using a composite objective function  consisting of a supervised component and an unsupervised component (an autoencoder loss). The objective function is listed below. More details can be found in \citet{shukla2019interpolationprediction}. In the fusion case, we use this objective to pre-train the interpolation-prediction network parameters in isolation.

\begin{align}
	\theta_*, \omega_* &=\argmin_{\theta, \omega} \sum_{n=1}^N \ell_P(y_n, g_{\omega}(f_{\theta}(\mbf{s}_n))) \nonumber\\
	& + \delta_R \sum_{n=1}^N \ell_I(\mbf{s}_n,\hat{\mbf{s}}_n) 
	 + \delta_F \Vert \theta \Vert_2^2 + \delta_G \Vert \omega \Vert_2^2 
%	\nonumber &\;\;\;\;\;\;\;\;\;\;\;\;+ \delta_R \sum_{n=1}^N \sum_{d=1}^D \sum_{j=1}^{L_{dn}} m_{jdn} \ell_I(x_{jdn},h_\theta^\chi(t_{jdn}, (1-\mbf{m}_n) \odot \mbf{s}_n))
\end{align}
where $\ell_P$ is the loss for the
prediction network and $\ell_I$ is the interpolation network autoencoder loss.

\begin{figure*}[t]
\begin{minipage}{0.5\linewidth}
\centering
\includegraphics[width=\linewidth]{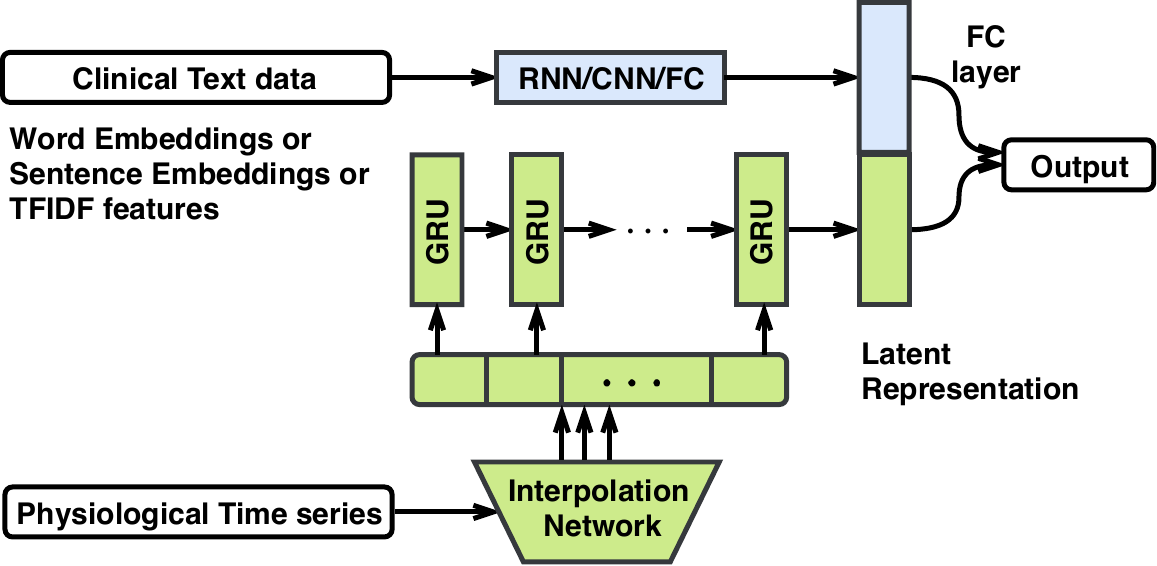}
%\caption{Late Fusion}
\end{minipage}
\begin{minipage}{0.5\linewidth}
\centering
\includegraphics[width=0.95\linewidth]{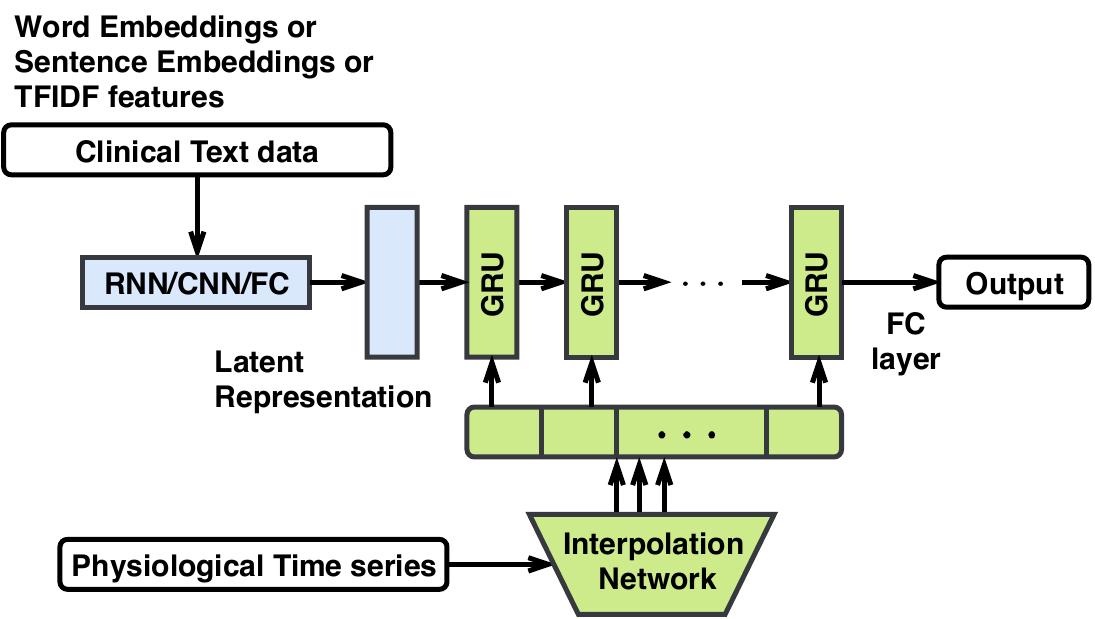}
%\caption{Early Fusion}
\end{minipage}
\caption{Architectures (left: late fusion, right: early fusion) for combining time series and text information}
\label{fig:text_vital}
\end{figure*}

\subsection{Text Models}
\label{sec:text}

We consider several different approaches to modeling unstructured text including approaches based on bag-of-words and word embedding representations. We describe each text representation approach below.

\begin{itemize}
\item {TF-IDF:} Each text document is first represented using TF-IDF features computed from a bag-of-words representation. We remove stop words and select a vocabulary consisting of the top 6,000 most frequent remaining words. We apply a one hidden layer (1NN) fully connected network of size 128 on top of the TF-IDF inputs, followed by the rest of the prediction network.

\item{Word Embedding (WE):} Each document is first represented as a matrix where rows are words in the document and columns are word embedding dimensions. Word embeddings are computed using a standard, pre-trained 300-dimensional GloVe model \citep{glove}. We then apply a convolutional neural network model with one 1D convolution and one pooling layer, followed by a fully-connected layer of size 128 connected to the rest of the prediction network. Stop words and words with no embeddings are removed. All documents are zero-padded to match the length of the longest document. We select the number of convolution kernels on a validation set. 

\item {Unweighted Sentence Embedding (USE):} Each document is first represented as a matrix where rows are sentences in the document and columns are the sentence embedding dimensions. The sentence embeddings are computed by averaging the GloVe embeddings \cite{glove} of their constituent words. We then apply a GRU model \cite{gru} to the sequence of sentence embeddings, followed by a fully-connected layer of size 128 connected to the rest of the prediction network. We consider GRU models with between 32 and 512 hidden units and select the best on a validation set.

%In this case, the temporal structure of text is preserved. 
%Recurrent models such as LSTM, GRU can be used on top to model these sentence embeddings. We use GRU in our experiments.

\item {Weighted Sentence Embedding (WSE):} Each document is represented as a matrix where rows are sentences in the document and columns are are the sentence embedding dimensions. We compute the sentence embedding by weighting the GloVe word embeddings \cite{glove} based on their unigram probability in the entire corpus as described in \citet{simple}. The remainder of this approach matches the unweighted case as described above.

%We compute the sentence embedding by weighing the word embeddings based on their unigram probability in the entire corpus as described in \cite{simple}. We use GRU on top to model the clinical text using the sentence embeddings. 
\end{itemize}

In all cases, the text representations described above are connected to the remainder of a prediction network via a 128-dimensional hidden layer. Recalling that $\mbf{v}_n$ represents the raw, unstructured text data available as input for data case $n$, we can view each of the methods described above as a different approach to computing
a fixed, 128-dimensional embedding $\hat{\mbf{v}}_n = h_\phi(\mbf{v}_n)$. To learn the parameters $\phi$ for each approach, we use a supervised pre-training approach. We directly connect the text embedding layer $\hat{\mbf{v}}_n$ to the prediction target, and minimize a prediction loss. In this work, we focus on in-hospital mortality prediction and use binary cross entropy as the loss function during pre-training. 

\subsection{Fusion Approaches}
% We handle heterogeneity between physiological time series data and clinical notes by learning fixed-length latent representations and propose fusion approaches on how to combine them. We learn a fixed-dimensional latent representation of the physiological time series data by applying a GRU layer on top of the interpolation network described in Section \ref{sec:interp}. Similarly, we learn a note embedding for the clinical notes data with the embedding-based models described in \ref{sec:text} which is again a fixed dimensional vector. We present two prediction network architectures for fusion that accept as input the interpolants produced by the interpolation network and the text embedding produced by the unstructured text models. Both architectures are shown in Figure \ref{fig:text_vital} and are described below.

In this section, we present fusion architectures that combine the interpolation-prediction network described in Section \ref{sec:interp} with the embedding-based models for representing unstructured text described in Section \ref{sec:text}. In particular, we present two prediction network architectures that accept as input the interpolants produced by the interpolation network and the text embeddings produced by the unstructured text models. Both architectures are shown in Figure \ref{fig:text_vital} and are described below.

\begin{itemize}
\item {\bf Late Fusion:} In this approach, the prediction network uses the same GRU architecture used by \citet{shukla2019interpolationprediction} to extract a fixed-dimensional latent representation of the physiological time series data. This representation is concatenated with the text embedding layer and the combined latent representation is connected to the prediction target using a linear layer. This architecture is shown in Figure \ref{fig:text_vital} (left).

\item {\bf Early Fusion:} We also consider a deeper integration of the information contained in both physiological time series and clinical notes. In this method, our prediction network has access to the clinical text data prior to incorporating physiological time series data via a GRU layer, as shown in Figure \ref{fig:text_vital} (right). 
%This method thus respects the timeline of observations as the clinical text data used in this work primarily comprise the chief complaint and past medical history, which are available before the physiological time series data are recorded. 

% \item {\bf Early Fusion:} We also consider a deeper integration of the information contained in both physiological time series and clinical notes. In this method, we learn the latent representation of the physiological time series data using interpolation network outputs in a similar way but in the presence of the note embedding and thus we treat this as a final representation of both physiological time series and clinical notes. Again, the final representation is connected to the prediction target using a linear layer. The early-fusion approach is shown in Figure \ref{fig:text_vital} (right).
\end{itemize}

In both fusion architectures, the prediction network takes as input the time series interpolants $\hat{\mathbf{s}}_{n}=f_{\theta}(\mbf{s}_n)$ (where $f_{\theta}$ denote the interpolation network) as well as the  text embedding $\hat{\mathbf{v}}_{n}=h_{\phi_*} (\mbf{v}_n)$ and outputs a prediction 
$\hat{y}_n=g_{\omega}(\hat{\mbf{s}}_n, \hat{\mbf{v}}_n)=
g_{\omega}(f_{\theta}(\mbf{s}_n), h_{\phi_*} (\mbf{v}_n))$.
As described above, we use supervised pre-training of all of the model parameters by training the interpolation-prediction and text embedding networks in isolation. During the fusion stage, we freeze the text embedding parameters $\phi$ to their optimal pre-trained values $\phi_*$, and fine-tune the interpolation and prediction network parameters $\theta$ and $\omega$. 

The learning objective for
the fusion framework requires specifying a loss $\ell_P$ for the prediction
network (we use cross-entropy loss for classification). We let $\ell_I$ be the interpolation network autoencoder loss as described in Section \ref{sec:interp}.
We also include $\ell_2$ regularizers for all the network parameters. $\delta_F$, $\delta_G$, and $\delta_R$ are
hyper-parameters that control the trade-off between the components of the
objective function. The full objective is shown below.

\begin{align}
	\theta_*,\omega_* &=\argmin_{\theta,\omega} \sum_{n=1}^N \ell_P(y_n, g_{\omega}(f_{\theta}(\mbf{s}_n), h_{\phi_*}(\mbf{v}_n))) \nonumber\\
	&+ \delta_R \sum_{n=1}^N \ell_I(\mbf{s}_n,\hat{\mbf{s}}_n) 
	+\delta_F \Vert \theta \Vert_2^2 + \delta_G \Vert \omega \Vert_2^2 
%	\nonumber &\;\;\;\;\;\;\;\;\;\;\;\;+ \delta_R \sum_{n=1}^N \sum_{d=1}^D \sum_{j=1}^{L_{dn}} m_{jdn} \ell_I(x_{jdn},h_\theta^\chi(t_{jdn}, (1-\mbf{m}_n) \odot \mbf{s}_n))
\end{align}
% \begin{align}
%     \label{eq:late}
%     g_{\omega}(f_{\theta}(\mbf{s}_n), h_{\phi_*}(\mbf{v}_n)) = G_{\omega_2}(G_{\omega_1}(f_{\theta}(\mbf{s}_n)), h_{\phi_*}(\mbf{v}_n)) \\
%     \label{eq:early}
%     g_{\omega}(f_{\theta}(\mbf{s}_n), h_{\phi_*}(\mbf{v}_n)) = G_{\omega_2}(G_{\omega_1}(f_{\theta}(\mbf{s}_n), h_{\phi_*}(\mbf{v}_n))) 
% \end{align}

% The prediction network $g_{\omega}$ for late and early fusion approach is defined using Equation \ref{eq:late} and \ref{eq:early} respectively, where $\omega = \{\omega_1, \omega_2\}$, $G_{\omega_1}$ and $G_{\omega_2}$ are GRU and linear layer respectively.

Note again that we leverage a  pre-trained text embedding model, thus the text embedding model parameters $\phi$ are fixed to their optimal pre-trained values $\phi_*$. The parameters of the fusion model (as well as all other models used in this work) are learned using the Adam optimization method in TensorFlow with gradients provided via automatic differentiation. 

%\subsection{Implementation Details}
%All models are learned using the Adam optimization method in TensorFlow with gradients
%provided via automatic differentiation. We use 300 dimensional Glove embeddings to 
%represent each word as a vector.  

%% file: experiments.tex
%!TEX root = amia.tex

\section{Experiments and Results}

In this section, we present experiments and results. Our experiments focus on the relative predictive performance of text-only models, time series-only models, and fusion models for the problem of in-hospital mortality prediction. The prediction output is a single binary variable representing the occurrence of in-hospital morality more than 48 hours after admission.
The time series inputs to the prediction task are sparse and irregularly sampled physiological time series. We consider making predictions using physiological time series data available between 6 and 48 hours after admission. The text inputs to the prediction task consists of text content known at the time of admission and progress notes available between 6 and 48 hours after admission.  
% also vary through time and include text content known at the time of admission, text content known up to and including the first 24 hours after admission, and text content known up to and including the first 48 hours after admission. 
We begin by briefly describing the data set used, followed by the set of baseline and comparison models,  the empirical protocols used, and finally the results. 

\begin{table}[h]
\centering
           \footnotesize
            %\begin{minipage}{.5\linewidth}
                \begin{tabular}[h]{l c}
                
                 \toprule
                 {feature} & {Sampling Rate}\\
                    \midrule
                    SpO2 &  $0.80$\\
                    HR  &  $0.90$\\
                    RR &  $0.48$\\
                    SBP &  $0.59$\\
                    DBP &  $0.60$\\
                    Temp &  $0.19$\\
  
                    \bottomrule
                    \end{tabular}  
                    \,\,\,\,\,\,
                   % \end{minipage} &
       %\begin{minipage}{.5\linewidth}
                \begin{tabular}[h]{l c}
                 \toprule
                 {feature}  & {Sampling Rate}\\
                    \midrule
                    TGCS &  $0.14$\\
                    CRR & $0.06$\\
                    UO & $0.20$\\
                    FiO2 &$0.06$\\
                    Glucose &$0.10$\\
                    pH &$0.04$\\
                    \bottomrule
                    \end{tabular}
                   % \end{minipage}
                   \caption{Features extracted from MIMIC III}
                   \label{table:mis}
                \end{table}

\subsection{Dataset}
Our experiments are based on the publicly available MIMIC-III dataset \citep{johnson2016mimic}. This data set contains sparse and irregularly sampled physiological signals, discharge summaries, progress notes, medications, diagnostic codes, in-hospital mortality, length of stay, demographics information and more. It consists of approximately 58,000 hospital admission records. We focus on predicting in-hospital mortality using both the clinical text and time series data. We start with the dataset\footnote{https://github.com/mlds-lab/interp-net} used in \citet{shukla2019interpolationprediction} which consists of hospital admission records with hospital admission-to-discharge length of stay more than 48 hours. From that dataset, we obtained 42,984 records for our experiments after removing newborns and hospital admission records containing no clinical notes. A hospital admission may correspond to zero or multiple ICU episodes. In this paper, we only consider the data cases that were admitted to ICU at least once during their hospital stay. 

Similar to \citet{shukla2019interpolationprediction}, we extract 12 standard physiological variables from each of the records.  Table \ref{table:mis} shows the variables and sampling rates (per hour).
We use text data known at the time of admission such as chief complaints, past medical history and history of present illness. We take care in extracting this information from discharge summaries in order to avoid any information leak. We also extract progress notes from non-discharge reports such as respiratory, ECG, echo, radiology, and nursing reports. We use the date and time stamps on these reports to create a set of notes available between 6 and 48 hours after admission. Note that the physiological data and clinical notes are aligned in a conservative manner. If a clinical note has both a date and time associated with it, we assume that information was available at the specified time. For notes that have dates but not times available, we assume that information was available at the end of the indicated day. The happens for some ECG and Echo reports in the data set.

%For some experiments, we use all notes available up to and including the first 24 hours after admission, as well as all notes available up to and including the first 48 hours after admission.

%\subsection{Prediction Task}
%In our experiments, each admission record corresponds to one data case $(\mbf{s_n},\mbf{v_n}, y_n)$. 
%Each data case $n$ consists of a sparse and irregularly sampled time series $\mbf{s}_n$ with $D=12$ 
%dimensions, and unstructured clinical text data $\mbf{v}_n$.  Each dimension $d$ 
%of $\mbf{s}_n$ corresponds to one of the $12$ vital sign time series mentioned in Table \ref{table:mis}. 
%$y_n$ is a binary indicator where $y_n=1$ indicates that the patient died at any point within the hospital 
%stay following the first 48 hours and $y_n=0$ indicates that the patient was discharged at any point after
% the first 48 hours. There are 3,623(8.43\%) patients with a $y_n=1$ mortality label. 
% The complete data set is $\mathcal{D}=\{(\mbf{s_n},\mbf{v_n},y_n)|n=1,...,N\}$, and there
%are $N=42,984$ data cases. The goal in the classification task is to learn a function $h$ of the
%form $\hat{y}_n \leftarrow h(\mbf{s}_n,\mbf{b}_n)$ where $\hat{y}_n$ is a discrete value.

\subsection{Baseline Models}
We compare fusion models with a number of baseline approaches that model the physiological 
time series or the clinical text data individually.  \citet{shukla2019interpolationprediction} show that the 
interpolation-prediction network outperforms 
a range of baseline and recently proposed models on both classification and regression tasks for 
sparse and irregularly sampled time series. Hence, we use interpolation-prediction networks as our 
time series-only baseline model. We use the pre-trained text-only models described in Section \ref{sec:text} to provide text-only baselines.

\subsection{Empirical Protocols}
% We first divide the dataset into standard train (80\%, 27510 data cases) and test (20\%, 8597 data cases) sets. We set aside another 20\% (6877 data cases) from the training set to use as a validation set. Regarding the data set split used, 
Each unique hospital admission-to-discharge episode for a patient is assigned a unique ID in the MIMIC-III data set. The data in each episode are treated as being independent. In the train-test split, we divide the data based on the hospital admission ID (i.e. 80\% (27510) of IDs are used for training and 20\% (8597) are used for testing). We set aside another 20\% (6877 data cases) from the training set to use as a validation set. Since we only use data from within individual hospital-to-discharge episodes, the data cases we construct are temporally non-overlapping. Again, this is consistent with how the MIMIC-III data set has been used in past research \citep{shukla2019interpolationprediction,che2016recurrent}.

All models are trained to minimize the cross entropy loss. For all of the models, we independently tune the hyper-parameters - number of hidden layers, hidden units, convolutional filters, filter-size, learning rate, dropout rates and regularization parameters on the validation set. For TF-IDF-based models, we also tune the number of TF-IDF features. The neural network models are learned using the Adam optimizer. 
Early stopping is used on the validation set. The final outputs of the hidden layers are used in a logistic layer that predicts the class. We evaluate all the 
models using an estimate of generalization performance computed on the test set. 
We report the performance on the test set in terms of the area under the ROC curve (AUC score).

\subsection{Results}
In this section, we present the results of the mortality prediction 
experiments. We begin with text-only and time series-only baseline results, followed by fusion model results.

\subsubsection{Text-Only Baselines:}
 Table \ref{table:text} shows the classification performance for the text-only models described in Section \ref{sec:text}. We evaluate all models in the case of text data available at the time of admission. These results show that the TF-IDF-based model performs significantly better than the embedding methods. This may be due to the fact that health-specific concepts are not well represented in the standard Glove embeddings used. Another possible reason could 
be the use of abbreviated terms, which are quite common in clinical notes. For this reason, we only consider the TF-IDF model when making predictions based on all the progress notes available after admission. We can see that prediction performance using the TF-IDF model increases significantly as more text data become available over time. 
\begin{table}
\centering
\begin{tabular}{ l c c c} 
 \toprule
{\bf Text Model} & {\bf Hours from}& {\bf AUC}   \\
& {\bf Admission} &\\
 \midrule
 WE / CNN      & 0 & $0.6974$\\
 WSE / RNN     & 0 & $0.7364$\\
 {USE} / RNN   & 0 & $0.7473$\\ 
 TF-IDF / 1-NN & 0 & $0.7965$\\
 TF-IDF / 1-NN & 6 & $0.8035$\\
 TF-IDF / 1-NN & 12 & $0.8173$\\
 TF-IDF / 1-NN & 18 & $0.8263$\\
 TF-IDF / 1-NN & 24 & $0.8410$\\
 TF-IDF / 1-NN & 30 & $0.8454$\\
 TF-IDF / 1-NN & 36 & $0.8503$\\
 TF-IDF / 1-NN & 42 & $0.8554$\\
 TF-IDF / 1-NN & 48 & $0.8627$\\
\bottomrule
 \end{tabular}
 \caption{Text-only baselines.}
 \label{table:text}
\end{table}

\subsubsection{Time Series-Only Baseline:} Table \ref{table:text_vitals} assesses 
the predictive performance of the time series-only interpolation-prediction network described in Section \ref{sec:interp}. As expected, predictive performance increases as the amount of observed physiological data increases. We note that the results reported here are different from that in \citet{shukla2019interpolationprediction} because of additional data filtering required for removing hospital admission records containing no clinical notes and neonates data. Comparing to the results in Table \ref{table:text}, we can see that the predictive value of the clinical text available at the time of admission exceeds that of the available physiological data until near the end of the 42 hour period following admission. The next set of experiment aims to assess whether these two modalities can result in improved performance when fused. 

% \begin{table}[h]
% \centering
% %\footnotesize
% \begin{tabular}{ c c } 
%  \toprule
%  {\bf Hours from} & {\bf AUC }    \\
%  {\bf Admission} \\
%  \midrule
% $	6	$	&	$	0.7106	$	\\
% $	12	$	&	$	0.7380	$	\\
% $	18	$	&	$	0.7645	$	\\
% $	24	$	&	$	0.7759	$	\\
% $	30	$	&	$	0.7902	$	\\
% $	36	$	&	$	0.7958	$	\\
% $	42	$	&	$	0.8023	$	\\
% $	48	$	&	$	0.8245	$	\\
% \bottomrule
%  \end{tabular}
%  \caption{Time series-only baseline evaluated  with  increasing  amount  of  physiological  signals.}
%  \label{table:vitals}
%  \end{table}

\begin{table}[h]
\centering
 \begin{tabular}{ c c c c } 
  \toprule
  {\bf Hours }& {\bf Time} & {\bf Early}  &  {\bf Late} \\
  {\bf from} & {\bf Series-Only} & {\bf Fusion} & {\bf Fusion}\\
  {\bf Admission} &  {\bf AUC }  & {\bf AUC } &  {\bf AUC }\\
  \midrule
%  $	6	$&$	0.7850	$&$	0.8027	$\\
%  $	12	$&$	0.7916	$&$	0.8161	$\\
%  $	18	$&$	0.8046	$&$	0.8138	$\\
%  $	24	$&$	0.8126	$&$	0.8256	$\\
%  $	30	$&$	0.8284	$&$	0.8324	$\\
%  $	36	$&$	0.8291	$&$	0.8320	$\\
%  $	42	$&$	0.8380	$&$	0.8376	$\\
%  $	48	$&$	0.8427	$&$	0.8453	$\\
$	6	$&$	0.7106	$&$	0.7850	$&$	0.8027	$\\
 $	12	$&$	0.7380	$&$	0.7916	$&$	0.8161	$\\
 $	18	$&$	0.7645	$&$	0.8046	$&$	0.8138	$\\
 $	24	$&$	0.7759	$&$	0.8126	$&$	0.8256	$\\
 $	30	$&$	0.7902	$&$	0.8284	$&$	0.8324	$\\
 $	36	$&$	0.7958	$&$	0.8291	$&$	0.8320	$\\
 $	42	$&$	0.8023	$&$	0.8380	$&$	0.8376	$\\
 $	48	$&$	0.8245	$&$	0.8427	$&$	0.8453	$\\
 \bottomrule
  \end{tabular}
  \caption{Performance of Time Series-Only baseline, Early and Late Fusion approach with notes available at admission time and increasing amount of physiological signals. }
 \label{table:text_vitals}
\end{table}

\subsubsection{Fusion Approaches:} 
\begin{figure}[t]
\centering
\includegraphics[width=0.98\linewidth]{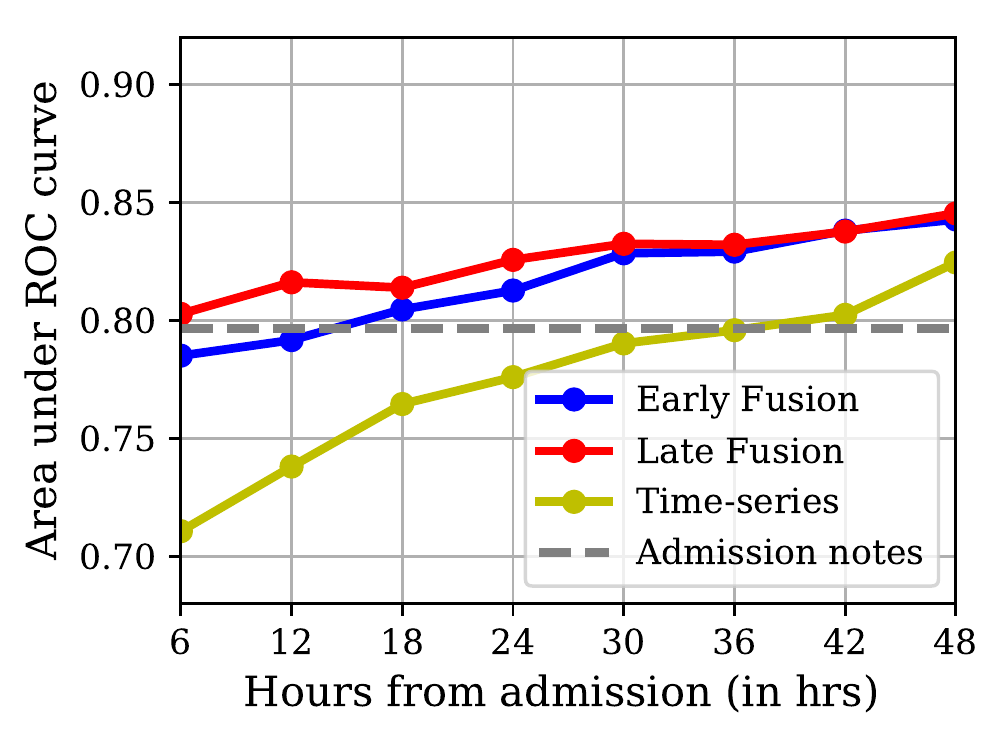}
\caption{Performance comparison on the mortality prediction task with
text available at admission only but increasing amounts of physiological time series.}
\label{fig:comp-admit}
\end{figure}

Based on the observed success of the TF-IDF-based model in the text-only baseline experiments, we examine the performance of fusion approaches using the TF-IDF-based model to embed the clinical text data. We begin by assessing the performance of a fusion approach that only has access to the clinical text data available at the time of admission, but increasing amounts of physiological time series data up to the end of the 48 hour period following admission. 
Table \ref{table:text_vitals} shows the classification performance of the early and late fusion models under this experimental scenario. Figure \ref{fig:comp-admit} shows the performance of early and late fusion relative to the time series-only and text-only baselines. We can see that the late fusion approach achieves better performance than the early fusion approach in the first 30 hours after admission, while both significantly improve on the time series-only baseline. However, we see that all three models that incorporate physiological data increase in predictive performance as the amount of physiological data increases. Further, we see that the performance gap between fusion and time series-only models decreases over time, which indicates that the advantage provided by the initial fusion with text data available at time of admission decreases over time as that information becomes less relevant. Finally, we note that the late fusion model outperforms the text-only baseline at all times while the early fusion model initially exhibits lower performance than the text-only TF-IDF baseline, but goes on to match and then outperforms the text-only baseline.

\begin{figure}[t]
\centering
\includegraphics[width=0.98\linewidth]{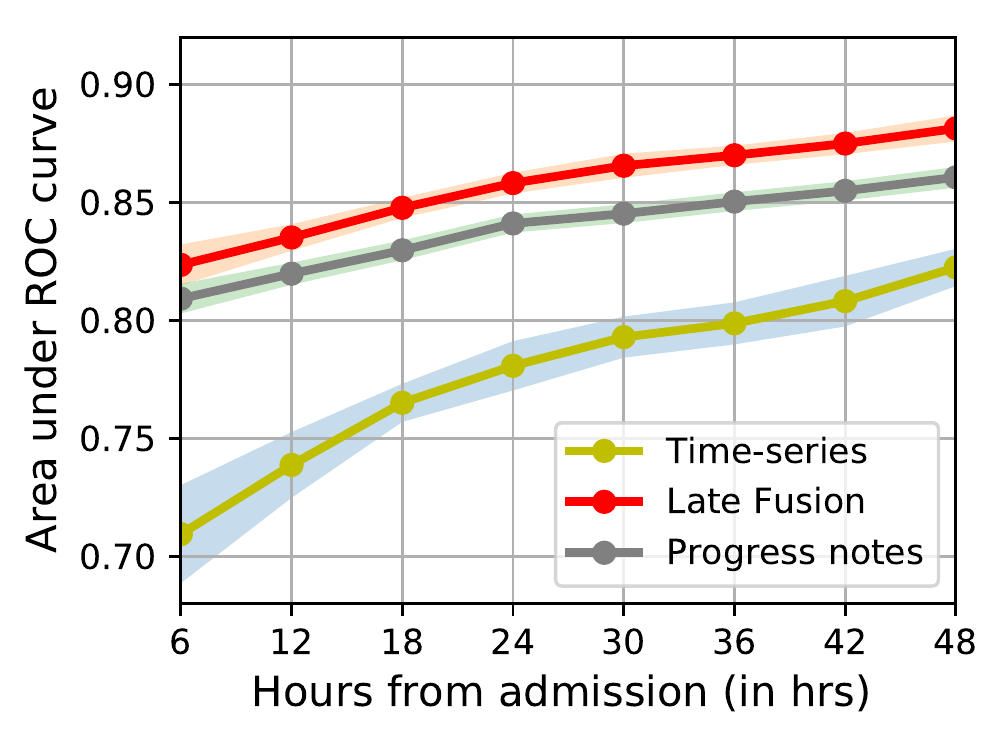}
\caption{Performance comparison on the mortality prediction task with increasing amount of physiological time series and progress notes using 5 randomly generated train-validation-test splits. Bands correspond to $95\%$ confidence interval around the mean.}
\label{fig:comp-rolling}
\end{figure}

Next, we consider the fusion process as increasing volumes of text data become available through time, as well as increasing volumes of physiological data. For this experiment, we consider only the TF-IDF-based text embedding model and limit the discussion to the late fusion approach as these models have achieved the best performance in our experiments to date. We consider incorporating text data known at admission at time 0, followed by the text of all notes known between  6  and  48  hours following admission . The results of this experiment are shown in Figure \ref{fig:comp-rolling}. We can see the predictive performance of progress notes is significantly better than physiological time series data. 
% We can see that there is a significant performance jump by incorporating notes available up to 24 hours, and a much smaller jump when incorporating notes available through the end of the full 48 hour period. 
We can also see that the performance of the fused model always exceeds that of the corresponding text-only baseline at a given point in time, with performance generally rising as additional physiological/text data become available. By comparing with Figure \ref{fig:comp-admit}, we can see that the addition of progress notes past admission results in a final fused model that significantly outperforms models that only have access to text data from the time of admission. 

To verify the statistical significance of the gap between the late fusion approach and the single-modality approaches, 
% shown in Figure \ref{fig:comp-rolling}, we conduct a follow-up experiment focusing on the case of 48 hours of data. 
we perform a five-fold random resampling assessment of the test AUC by randomly generating 5 train-validation-test splits. We run the complete hyper-parameter selection and learning pipeline for each approach on each of the five data sets. Figure \ref{fig:comp-rolling} shows the mean with $95\%$ confidence interval for all the baselines.    
% Table \ref{table:random_sampling} shows a summary of the  results. 
Further, we performed a paired t-test on the resulting collection of AUC values for the late fusion approach compared to the single modality approaches. The results show that the improvement in mean AUC shown in Figure \ref{fig:comp-rolling} is highly statistically significant $(p < 0.001)$. 

%\begin{table}[h]
%\caption{Classification performance with joint modeling of physiological time series and clinical text data}
%\begin{center}
%\begin{tabular}{ c c c  c c } 
% \toprule
% {\bf Hours from } & \multicolumn{2}{c}{\bf Early Fusion}  &  \multicolumn{2}{c}{\bf Late Fusion } \\
% {\bf Admission} &  {\bf AUC } & {\bf AUPRC } & {\bf AUC } & {\bf AUPRC } \\
% \midrule
%$	6	$&$	0.785	$&$ 0.2735 $&$	0.8027	$&$	0.3048 $\\
%$	12	$&$	0.7916	$&$ 0.2942 $&$	0.8161	$&$	0.3252$\\
%$	18	$&$	0.8046	$&$ 0.3008 $&$	0.8138	$&$	0.3317$\\
%$	24	$&$	0.8126	$&$ 0.3129 $&$	0.8256	$&$	0.3489$\\
%$	30	$&$	0.8284	$&$ 0.3523 $&$	0.8324	$&$	0.3654$\\
%$	36	$&$	0.8291	$&$ 0.3298 $&$	0.832	$&$	0.3710$\\
%$	42	$&$	0.838	$&$ 0.3851 $&$	0.8376	$&$	0.3815$\\
%$	48	$&$	0.8427	$&$ 0.4074 $&$	0.8453	$&$	0.4176$\\
%\bottomrule
% \end{tabular}
%\end{center}
%\label{table:text_vitals}
%\end{table} 
%
%\begin{figure}[h]
%\centering
%\includegraphics[width=0.6\linewidth]{Figures/compare4.pdf}
%\caption{Performance comparison on mortality prediction task}
%\label{fig:comp}
%\end{figure}

%\vspace{5mm}

% \begin{table}
% \centering
%  \begin{tabular}{ c c } 
%   \toprule
%   {\bf Model} &  {{\bf AUC} (mean $\pm$ std)}  \\
%   \midrule
%  Time-series only & $0.8224 \pm 0.0090$\\
%  Progress notes only & $0.8606 \pm 0.0050$\\
%  {\bf Late Fusion} &  $\mbf{0.8813 \pm 0.0063}$\\
%  \bottomrule
%   \end{tabular}
%   \caption{Performance comparison on the mortality prediction task using first 48 hours of data.}
%  \label{table:random_sampling}
% \end{table}

%% file: conclusions.tex
%!TEX root = amia.tex

\section{Discussion and Conclusions}

In this paper, we have developed methods for investigating the relative predictive value of the content of clinical notes and physiological time series data in ICU EHRs. We have considered models based on clinical text only, models based on physiological time-series only, and a novel fusion approach that combines both modalities. Our experiments have focused on using this methodology to assess the relative predictive value of clinical text and physiological data as a function of time since admission. We have focused on the task of predicting in-hospital mortality events which take place more than 48 hours after admission. 

Our results show that the relative value of information in text records known at the time of admission decreases over time as more physiological data are observed. However, incorporating newly available text data can significantly boost predictive performance. Finally, our results strongly support the conclusion that fusing both data modalities result in the best overall predictive performance.

%% file: main.bbl
\begin{thebibliography}{}

\bibitem[\protect\citeauthoryear{Arora, Liang, and Ma}{2017}]{simple}
Arora, S.; Liang, Y.; and Ma, T.
\newblock 2017.
\newblock A simple but tough-to-beat baseline for sentence embeddings.

\bibitem[\protect\citeauthoryear{Boag \bgroup et al\mbox.\egroup
  }{2015}]{cliner}
Boag, W.; Wacome, K.; Naumann, T.; and Rumshisky, A.
\newblock 2015.
\newblock Cliner : A lightweight tool for clinical named entity recognition.

\bibitem[\protect\citeauthoryear{Boag \bgroup et al\mbox.\egroup }{2018}]{boug}
Boag, W.; Doss, D.; Naumann, T.; and Szolovits, P.
\newblock 2018.
\newblock What's in a note? unpacking predictive value in clinical note
  representations.
\newblock {\em AMIA Joint Summits on Translational Science proceedings.}

\bibitem[\protect\citeauthoryear{Che \bgroup et al\mbox.\egroup
  }{2018}]{che2016recurrent}
Che, Z.; Purushotham, S.; Cho, K.; Sontag, D.; and Liu, Y.
\newblock 2018.
\newblock Recurrent neural networks for multivariate time series with missing
  values.
\newblock {\em Scientific Reports} 8(1):6085.

\bibitem[\protect\citeauthoryear{Choi, Yi-I~Chiu, and
  Sontag}{2016}]{choiconcept}
Choi, Y.; Yi-I~Chiu, C.; and Sontag, D.
\newblock 2016.
\newblock Learning low-dimensional representations of medical concepts.
\newblock {\em AMIA Joint Summits on Translational Science proceedings.}
  2016:41--50.

\bibitem[\protect\citeauthoryear{Chung \bgroup et al\mbox.\egroup }{2014}]{gru}
Chung, J.; G{\"{u}}l{\c c}ehre, {\c C}.; Cho, K.; and Bengio, Y.
\newblock 2014.
\newblock Empirical evaluation of gated recurrent neural networks on sequence
  modeling.
\newblock {\em arXiv e-prints} abs/1412.3555.

\bibitem[\protect\citeauthoryear{De~Vine \bgroup et al\mbox.\egroup
  }{2014}]{devine}
De~Vine, L.; Zuccon, G.; Koopman, B.; Sitbon, L.; and Bruza, P.
\newblock 2014.
\newblock Medical semantic similarity with a neural language model.
\newblock In {\em Proceedings of the 23rd ACM International Conference on
  Conference on Information and Knowledge Management}, CIKM '14,  1819--1822.

\bibitem[\protect\citeauthoryear{Fiterau \bgroup et al\mbox.\egroup
  }{2017}]{shortfuse}
Fiterau, M.; Bhooshan, S.; Fries, J.~A.; Bournhonesque, C.; Hicks, J.~L.;
  Halilaj, E.; R{\'e}, C.; and Delp, S.~L.
\newblock 2017.
\newblock Shortfuse: Biomedical time series representations in the presence of
  structured information.
\newblock In {\em MLHC}.

\bibitem[\protect\citeauthoryear{Futoma \bgroup et al\mbox.\egroup
  }{2017}]{futoma2017improved}
Futoma, J.; Hariharan, S.; Heller, K.; Sendak, M.; Brajer, N.; Clement, M.;
  Bedoya, A.; and O’Brien, C.
\newblock 2017.
\newblock An improved multi-output gaussian process rnn with real-time
  validation for early sepsis detection.
\newblock In {\em Machine Learning for Healthcare Conference},  243--254.

\bibitem[\protect\citeauthoryear{Ghassemi \bgroup et al\mbox.\egroup
  }{2012}]{ghasemmi}
Ghassemi, M.; Naumann, T.; Joshi, R.; and Rumshisky, A.
\newblock 2012.
\newblock Topic models for mortality modeling in intensive care units.
\newblock {\em ICML machine learning for clinical data analysis workshop,
  June}.

\bibitem[\protect\citeauthoryear{Ghassemi \bgroup et al\mbox.\egroup
  }{2015}]{ghassemi15}
Ghassemi, M.; Pimentel, M. A.~F.; Naumann, T.; Brennan, T.; Clifton, D.~A.;
  Szolovits, P.; and Feng, M.
\newblock 2015.
\newblock A multivariate timeseries modeling approach to severity of illness
  assessment and forecasting in icu with sparse, heterogeneous clinical data.
\newblock In {\em Proceedings of the Twenty-Ninth AAAI Conference on Artificial
  Intelligence}, AAAI'15,  446--453.

\bibitem[\protect\citeauthoryear{Hardoon, Szedmak, and
  Shawe-taylor}{2004}]{hardoon}
Hardoon, D.~R.; Szedmak, S.~R.; and Shawe-taylor, J.~R.
\newblock 2004.
\newblock Canonical correlation analysis: An overview with application to
  learning methods.
\newblock {\em Neural Comput.} 16(12):2639--2664.

\bibitem[\protect\citeauthoryear{Hotelling}{1935}]{harold}
Hotelling, H.
\newblock 1935.
\newblock Relations between two sets of variates.
\newblock {\em Biometrika} 28:321--377.

\bibitem[\protect\citeauthoryear{Jin \bgroup et al\mbox.\egroup
  }{2018}]{mlforh}
Jin, M.; Bahadori, M.~T.; Colak, A.; Bhatia, P.; Celikkaya, B.; Bhakta, R.;
  Senthivel, S.; Khalilia, M.; Navarro, D.; Zhang, B.; Doman, T.; Ravi, A.;
  Liger, M.; and Kass{-}Hout, T.~A.
\newblock 2018.
\newblock Improving hospital mortality prediction with medical named entities
  and multimodal learning.
\newblock {\em CoRR} abs/1811.12276.

\bibitem[\protect\citeauthoryear{Johnson \bgroup et al\mbox.\egroup
  }{2016}]{johnson2016mimic}
Johnson, A.~E.; Pollard, T.~J.; Shen, L.; Li-wei, H.~L.; Feng, M.; Ghassemi,
  M.; Moody, B.; Szolovits, P.; Celi, L.~A.; and Mark, R.~G.
\newblock 2016.
\newblock Mimic-iii, a freely accessible critical care database.
\newblock {\em Scientific data} 3:160035.

\bibitem[\protect\citeauthoryear{Kalchbrenner, Grefenstette, and
  Blunsom}{2014}]{blunsom}
Kalchbrenner, N.; Grefenstette, E.; and Blunsom, P.
\newblock 2014.
\newblock A convolutional neural network for modelling sentences.
\newblock In {\em ACL}.

\bibitem[\protect\citeauthoryear{Karpathy, Joulin, and
  Fei-Fei}{2014}]{karpathy14}
Karpathy, A.; Joulin, A.; and Fei-Fei, L.
\newblock 2014.
\newblock Deep fragment embeddings for bidirectional image sentence mapping.
\newblock In {\em Proceedings of the 27th International Conference on Neural
  Information Processing Systems},  1889--1897.

\bibitem[\protect\citeauthoryear{Kiela and Bottou}{2014}]{kiela}
Kiela, D., and Bottou, L.
\newblock 2014.
\newblock Learning image embeddings using convolutional neural networks for
  improved multi-modal semantics.
\newblock In {\em EMNLP}.

\bibitem[\protect\citeauthoryear{Klein \bgroup et al\mbox.\egroup
  }{2015}]{klein}
Klein, B.; Lev, G.; Sadeh, G.; and Wolf, L.
\newblock 2015.
\newblock Associating neural word embeddings with deep image representations
  using fisher vectors.
\newblock  4437--4446.

\bibitem[\protect\citeauthoryear{Lehman \bgroup et al\mbox.\egroup
  }{2012}]{lehman}
Lehman, L.-w.; Saeed, M.; Long, W.; Lee, J.; and Mark, R.
\newblock 2012.
\newblock Risk stratification of icu patients using topic models inferred from
  unstructured progress notes.
\newblock {\em AMIA Annual Symposium proceedings} 2012:505--11.

\bibitem[\protect\citeauthoryear{Li and Marlin}{2015}]{li2015classification}
Li, S. C.-X., and Marlin, B.~M.
\newblock 2015.
\newblock Classification of sparse and irregularly sampled time series with
  mixtures of expected {G}aussian kernels and random features.
\newblock In {\em 31st Conference on Uncertainty in Artificial Intelligence}.

\bibitem[\protect\citeauthoryear{Li and Marlin}{2016}]{li2016scalable}
Li, S. C.-X., and Marlin, B.~M.
\newblock 2016.
\newblock A scalable end-to-end gaussian process adapter for irregularly
  sampled time series classification.
\newblock In {\em Advances In Neural Information Processing Systems},
  1804--1812.

\bibitem[\protect\citeauthoryear{Lipton, Kale, and
  Wetzel}{2016}]{lipton2016directly}
Lipton, Z.~C.; Kale, D.; and Wetzel, R.
\newblock 2016.
\newblock Directly modeling missing data in sequences with rnns: Improved
  classification of clinical time series.
\newblock In {\em Machine Learning for Healthcare Conference},  253--270.

\bibitem[\protect\citeauthoryear{Little and
  Rubin}{2014}]{little2014statistical}
Little, R.~J., and Rubin, D.~B.
\newblock 2014.
\newblock {\em Statistical analysis with missing data}, volume 333.

\bibitem[\protect\citeauthoryear{Lu \bgroup et al\mbox.\egroup }{2008}]{Lu2008}
Lu, Z.; Leen, T.~K.; Huang, Y.; and Erdogmus, D.
\newblock 2008.
\newblock A reproducing kernel hilbert space framework for pairwise time series
  distances.
\newblock In {\em Proceedings of the 25th International Conference on Machine
  Learning}, ICML '08,  624--631.

\bibitem[\protect\citeauthoryear{Marlin \bgroup et al\mbox.\egroup
  }{2012}]{marlin-ihi2012}
Marlin, B.~M.; Kale, D.~C.; Khemani, R.~G.; and Wetzel, R.~C.
\newblock 2012.
\newblock Unsupervised pattern discovery in electronic health care data using
  probabilistic clustering models.
\newblock In {\em Proceedings of the 2nd ACM SIGHIT International Health
  Informatics Symposium},  389--398.

\bibitem[\protect\citeauthoryear{Minarro~Gimenez, Marin~Alonso, and
  Samwald}{2014}]{minarro}
Minarro~Gimenez, J.~A.; Marin~Alonso, O.; and Samwald, M.
\newblock 2014.
\newblock Exploring the application of deep learning techniques on medical text
  corpora.
\newblock {\em Studies in health technology and informatics} 205:584--588.

\bibitem[\protect\citeauthoryear{Ngiam \bgroup et al\mbox.\egroup
  }{2011}]{ngiam}
Ngiam, J.; Khosla, A.; Kim, M.; Nam, J.; Lee, H.; and Ng, A.~Y.
\newblock 2011.
\newblock Multimodal deep learning.
\newblock In {\em ICML 2011}.

\bibitem[\protect\citeauthoryear{Pennington, Socher, and Manning}{2014}]{glove}
Pennington, J.; Socher, R.; and Manning, C.~D.
\newblock 2014.
\newblock Glove: Global vectors for word representation.
\newblock In {\em In EMNLP}.

\bibitem[\protect\citeauthoryear{Rajkomar \bgroup et al\mbox.\egroup
  }{2018}]{rajkomar_multi}
Rajkomar, A.; Oren, E.; Chen, K.; Dai, A.~M.; Hajaj, N.; Hardt, M.; Liu, P.~J.;
  Liu, X.; Marcus, J.; Sun, M.; Sundberg, P.; Yee, H.; Zhang, K.; Zhang, Y.;
  Flores, G.; Duggan, G.~E.; Irvine, J.; Le, Q.; Litsch, K.; Mossin, A.;
  Tansuwan, J.; Wang, D.; Wexler, J.; Wilson, J.; Ludwig, D.; Volchenboum,
  S.~L.; Chou, K.; Pearson, M.; Madabushi, S.; Shah, N.~H.; Butte, A.~J.;
  Howell, M.~D.; Cui, C.; Corrado, G.~S.; and Dean, J.
\newblock 2018.
\newblock Scalable and accurate deep learning with electronic health records.
\newblock {\em npj Digital Medicine} 1(1):18.

\bibitem[\protect\citeauthoryear{Rodrigues, Markou, and
  Pereira}{2018}]{rodrigues}
Rodrigues, F.; Markou, I.; and Pereira, F.
\newblock 2018.
\newblock Combining time-series and textual data for taxi demand prediction in
  event areas: A deep learning approach.
\newblock {\em Information Fusion} 49.

\bibitem[\protect\citeauthoryear{Savova \bgroup et al\mbox.\egroup
  }{2010}]{jamia2}
Savova, G.; Masanz, J.; V~Ogren, P.; Zheng, J.; Sohn, S.; C~Kipper-Schuler, K.;
  and Chute, C.
\newblock 2010.
\newblock Mayo clinical text analysis and knowledge extraction system (ctakes):
  Architecture, component evaluation and applications.
\newblock {\em Journal of the American Medical Informatics Association : JAMIA}
  17:507--13.

\bibitem[\protect\citeauthoryear{Shukla and
  Marlin}{2019}]{shukla2019interpolationprediction}
Shukla, S.~N., and Marlin, B.
\newblock 2019.
\newblock Interpolation-prediction networks for irregularly sampled time
  series.
\newblock In {\em International Conference on Learning Representations}.

\bibitem[\protect\citeauthoryear{Silberer and Lapata}{2014}]{silberer}
Silberer, C., and Lapata, M.
\newblock 2014.
\newblock Learning grounded meaning representations with autoencoders.
\newblock In {\em ACL}.

\bibitem[\protect\citeauthoryear{Socher \bgroup et al\mbox.\egroup
  }{2014}]{socher14}
Socher, R.; Karpathy, A.; Le, Q.~V.; Manning, C.~D.; and Ng, A.~Y.
\newblock 2014.
\newblock Grounded compositional semantics for finding and describing images
  with sentences.
\newblock {\em Transactions of the Association for Computational Linguistics}
  2:207--218.

\bibitem[\protect\citeauthoryear{Srivastava and
  Salakhutdinov}{2012}]{srivastavamulti}
Srivastava, N., and Salakhutdinov, R.~R.
\newblock 2012.
\newblock Multimodal learning with deep boltzmann machines.
\newblock In Pereira, F.; Burges, C. J.~C.; Bottou, L.; and Weinberger, K.~Q.,
  eds., {\em Advances in Neural Information Processing Systems 25}.
\newblock  2222--2230.

\bibitem[\protect\citeauthoryear{Tang, Yang, and Zhou}{2009}]{tang}
Tang, X.; Yang, C.; and Zhou, J.
\newblock 2009.
\newblock Stock price forecasting by combining news mining and time series
  analysis.
\newblock In {\em Proceedings of the 2009 IEEE/WIC/ACM International Joint
  Conference on Web Intelligence and Intelligent Agent Technology - Volume 01},
  WI-IAT '09,  279--282.

\bibitem[\protect\citeauthoryear{Uzuner, Solti, and Cadag}{2010}]{jamia3}
Uzuner, O.; Solti, I.; and Cadag, E.
\newblock 2010.
\newblock Extracting medication information from clinical text.
\newblock {\em Journal of the American Medical Informatics Association : JAMIA}
  17:514--518.

\bibitem[\protect\citeauthoryear{Xu \bgroup et al\mbox.\egroup
  }{2018}]{xu_multi}
Xu, Y.; Biswal, S.; Deshpande, S.~R.; Maher, K.~O.; and Sun, J.
\newblock 2018.
\newblock Raim: Recurrent attentive and intensive model of multimodal patient
  monitoring data.
\newblock In {\em Proceedings of the 24th ACM SIGKDD International Conference
  on Knowledge Discovery \& Data Mining}, KDD ?18,  2565?2573.
\newblock New York, NY, USA: Association for Computing Machinery.

\bibitem[\protect\citeauthoryear{Yoon, Zame, and van~der
  Schaar}{2017}]{Yoon_mRNN}
Yoon, J.; Zame, W.~R.; and van~der Schaar, M.
\newblock 2017.
\newblock Multi-directional recurrent neural networks : A novel method for
  estimating missing data.

\end{thebibliography}
